\documentclass{article}

    \PassOptionsToPackage{numbers}{natbib}

\usepackage[preprint]{neurips_2021}

\usepackage[utf8]{inputenc} 
\usepackage[T1]{fontenc}    
\usepackage{hyperref}       
\usepackage{url}            
\usepackage{booktabs}       
\usepackage{amsfonts}       
\usepackage{nicefrac}       
\usepackage{microtype}      
\usepackage{xcolor}         
\usepackage{amsmath}
\usepackage{amsthm}
\usepackage{graphicx}

\usepackage{bbm}
\usepackage{placeins}
\usepackage{subfigure}

\newcommand{\ie}{\emph{i.e.}}
\newcommand{\eg}{\emph{e.g.}}

\newcommand{\real}{\mathbb{R}}

\title{Contrastive GAN}
\title{Contrastive Generative Adversarial Networks}
\title{Training Generative Adversarial Networks via Approximate Kernel Density Estimation}
\title{Kernel Density Estimation-Based Generative Adversarial Networks}
\title{Generative Adversarial Learning via\\Kernel Density Discrimination}

\author{
  Abdelhak Lemkhenter \\
  Institute of Computer Science\\
  University of Bern\\
  \texttt{abdelhak.lemkhenter@inf.unibe.ch} \\
   \And
   Adam Bielski \\
    Institute of Computer Science\\
  University of Bern\\
  \texttt{adam.bielski@inf.unibe.ch} \\
   \AND
   Alp Eren Sari\\
    Institute of Computer Science\\
  University of Bern\\
  \texttt{alp.sari@inf.unibe.ch} \\
   \And
   Paolo Favaro \\
    Institute of Computer Science\\
  University of Bern\\
  \texttt{paolo.favaro@inf.unibe.ch} \\
}

\begin{document}

\maketitle

\begin{abstract}
We introduce Kernel Density Discrimination GAN (KDD~GAN), a novel method for generative adversarial learning. KDD~GAN formulates the training as a likelihood ratio optimization problem where the data distributions are written explicitly via (local) Kernel Density Estimates (KDE). This is inspired by the recent progress in contrastive learning and its relation to KDE.
We define the KDEs directly in feature space and forgo the requirement of invertibility of the kernel feature mappings. 
In our approach, features are no longer optimized for linear separability, as in the original GAN formulation, but for the more general discrimination of distributions in the feature space. 
We analyze the gradient of our loss with respect to the feature representation and show that it is better behaved than that of the original hinge loss. We perform experiments with the proposed KDE-based loss, used either as a training loss or a regularization term, on both CIFAR10 and scaled versions of ImageNet. 
We use BigGAN/SA-GAN as a backbone and baseline, since our focus is not to design the architecture of the networks. We show a boost in the quality of generated samples with respect to FID from 10\% to 40\% compared to the baseline. Code will be made available.
\end{abstract}

\section{Introduction}

Generative learning finds applications in many computer vision applications such as image translation \cite{isola2017image, zhu2017unpaired, dundar2020panoptic, park2020contrastive}, image processing \cite{ledig2017photo, kupyn2018deblurgan}, image restoration \cite{ulyanov2018deep, pan2020exploiting, zhang2016colorful}, text to image mapping \cite{reed2016generative, zhang2017stackgan, li2020manigan, ramesh2021zero} and, more in general, to define image priors in image-based optimization problems \cite{ulyanov2018deep, menon2020pulse}.
Generative models based on adversarial learning have been widely successful thanks to several breakthroughs in the design of the generator and discriminator architectures \cite{brock2018large, zhang2019self, karras2019style}, of the loss functions \cite{arjovsky2017wasserstein, kang2020contragan, yu2021dual} and regularization methods \cite{mao2017least,  miyato2018spectral, zhang2019consistency, kang2020contragan}.
Yet, the training of generative models is not straightforward and can be still prone to mode collapse \cite{srivastava2017veegan, yu2020inclusive, liu2021divco} or the inability to capture long-range statistics in the data, which leads to visible artifacts \cite{zhang2019self, lin2021infinitygan}.

One key assumption in the basic formulation of adversarial learning of \cite{goodfellow2014generative} is that the generator network should compete with an optimal discriminator, that is, a classifier that can tell  real from generated data apart if any of their statistics does not match. Thus, the general wisdom is that the more powerful the discriminator is and the better the generator trains. 
Given that training models with contrastive losses yields better performance than training with cross-entropy losses  \cite{Khosla2020}, and that contrastive learning can be seen as introducing Kernel Density Estimate (KDE) approximations of the data distribution \cite{wang2020understanding}, we propose to train the discriminative and generative models through a KDE approximation of the likelihood ratio loss.
Moreover, this approach ensures that the loss defines a valid statistical divergence between the distributions of the real and generated data at all times. In contrast, the loss used to train state of the art generative adversarial networks corresponds to a known statistical divergence between distributions of real and fake data only when at the saddle point of the min-max game. 
Our analysis shows that the gradients of the proposed loss are better behaved than those of the hinge loss (as defined, for example, by \cite{miyato2018cgans}).
We propose a KDE defined directly in feature space, so that non-invertible features are allowed. Our method includes a much broader set of discriminator solutions than in the binary classification task of the original GAN formulation. In fact, in the KDE approach the features are no longer optimized for linear separability, but for the more general discrimination of distributions in the feature space. We call our method \emph{Kernel Density Discrimination GAN} (KDD~GAN).

The recent method of \cite{kang2020contragan}, ContraGAN, also proposes contrastive losses for GAN training. However, their approach is fundamentally different from the formulation in this paper. One key aspect of ContraGAN is that the contrastive loss is explicitly used as a regularizer, while our formulation can be used also on its own. A second important aspect is that the contrastive formulation of ContraGAN is used to explicitly separate the features of different categories, but this is not the case in our formulation. In fact, our approach can be used either for the conditional or unconditional training. 

\noindent\textbf{Contributions}: 1) KDD~GAN achieves an improvement of more than $10\%$ in the FID and IS performance compared to the baseline (BigGAN of \cite{brock2018large}) on CIFAR10 \cite{krizhevsky2009learning} and Tiny ImageNet \cite{Le2015TinyIN}; 
2) The KDD loss can also be used as a regularizer and improve the training in terms of FID and IS in CIFAR10, Tiny ImageNet, and ImageNet $64\times 64$, which has images scaled to $64\times 64$ pixels (derived from \cite{deng2009imagenet}); 3) The proposed KDD loss is flexible and can be combined with other methods; moreover, our implementation does not increase the computational load and the memory usage excessively compared to the conventional hinge loss training  \cite{miyato2018cgans}.



\section{Kernel Density Discrimination}\label{sec:kdd}

Let $S_r = \{x_r^{(1)},\dots, x_r^{(m)}\}$ be a dataset of $m$ image samples $x_r^{(i)}\in\real^d$, which we call \emph{real data}. They are the instances of a probability density function (pdf) $p_r$, which we call the \emph{real data} pdf. We aim to build a generative model that maps zero-mean Gaussian samples to images, and such that they also follow the real data distribution. To distinguish the generated samples, we denote the dataset of generated data by $S_g$, a generated image sample by $x_g$, and the \emph{generated data} pdf by $p_g$.

We build our generative model through adversarial learning as in the pioneering work of \cite{goodfellow2014generative}, and thus work with a \emph{discriminator} network $D$ and a \emph{generator} network $G$. Then, generative adversarial learning can be cast as the following bilevel optimization problem 
\begin{align}
\centering
    \min_{G}~\mathcal{L}_G(D^\ast,G)\label{eq:bilevel}\quad \text{such that }~D^\ast = \arg\min_D \mathcal{L}_D(D,G),
\end{align}
where the optimization in $G$ and $D$ is implemented as the optimization with respect to the parameters of the neural networks implementing $G$ and $D$.
In the case of hinge loss optimization (see \eg, \cite{miyato2018cgans}), the losses in eq.~\eqref{eq:bilevel} are defined as 
\begin{align}
\textstyle
    \mathcal {L}_D(D,G) = \frac{1}{|S_g|} \sum_{x_g\in S_g} \max\left\{0,1+D(x_g)\right\} +
    \frac{1}{|S_r|}\sum_{x_r\in S_r} \max\left\{0,1-D(x_r)\right\}
\end{align}
and $\mathcal{L}_G(D^\ast,G) = \nicefrac{1}{|S_g|}\sum_{x_g \in S_g} -D^\ast(x_g)$,
which rely on the assumption that the discriminator takes the form of $D(x) = \log p_r(x) - \log p_g(x)$.
In our approach, we would like instead to explicitly approximate the form $\log p_r(x) - \log p_g(x)$. 
The main advantage of having this form is that it is a well-defined divergence between distributions. Thus, it defines a valid gradient for the generator at all times, modulo the errors due to the approximation. 

One option is to approximate $p_r(x)$ and $p_g(x)$ with Kernel Density Estimates (KDE). The kernels could be defined in some feature space, and the feature mapping could be estimated during training. However, one restriction is that, in general, it is necessary to guarantee the invertibility of the feature map. This is the same requirement of Normalizing Flows (see, \eg, \cite{kobyzev2020normalizing}) and thus one would have to follow similar restrictions in the neural architectures used to compute the features. Moreover, the training invertible neural networks is not easy. Indeed, despite their mathematically sound definition, Normalizing Flows are not yet on par with other generative learning methods. 
To simplify the training of the generative model, we propose instead to use KDEs in feature space, defined by $\phi(x) = D(x)\in\real^K$, and to allow the feature mapping to be non-invertible. Thus, we aim to match the push-forward measures $\phi_\ast p_r$ and $\phi_\ast p_g$, which we denote by $\hat p_r^\phi$ and $\hat p_g^\phi$ respectively. If the divergence between the push-forward measures is minimized, and the corresponding generated data pdf $p_g$ has a different support than $p_r$, then we can use Theorem~2.1 and 2.2 of \cite{arjovsky2017towards} to show a contradiction (\ie, that there exists an optimal discriminator that separates the supports of $p_g$ and $p_r$, and thus such that the match in feature space cannot be exact). This implies that matching the push-forward measures for any feature $\phi$ ensures that the supports of $p_g$ and $p_r$ also match. 
We write the losses explicitly as
\begin{align}
    \mathcal {L}_D(\phi,G) = \sum_{x_g\in S_g} 
    \frac{\max\left\{0,1+\log \frac{\hat p_{r}^\phi(\phi(x_g))}{\hat p_{g}^\phi(\phi(x_g))}\right\}}{|S_g|}
     +
    \sum_{x_r\in S_r}
    \frac{\max\left\{0,1-\log \frac{\hat p_{r}^\phi(\phi(x_r))}{\hat p_{g}^\phi(\phi(x_r))}\right\}}{|S_r|}
    \label{eq:LDunsupervised}
\end{align}
and $\mathcal{L}_G(\phi^\ast,G) = \nicefrac{1}{|S_g|}\sum_{x_g \in S_g} -\log \frac{\hat p_{r}^{\phi^\ast}(\phi^\ast(x_g))}{\hat p_{g}^{\phi^\ast}(\phi^\ast(x_g))}$,
by approximating the push-forward measures of the pdfs $p_r$ and $p_g$ via the following KDEs in feature space \begin{alignat}{5}
    \hat p_{r}^{\phi}(\xi) &= \frac{1}{|S_r|}\sum_{x_r\in S_r} \mathcal{K}_{\tau}(\phi(x_r),\xi),&&\quad\quad\quad
    &\hat p_{g}^{\phi}(\xi) &= \frac{1}{|S_g|}\sum_{x_g\in S_g} \mathcal{K}_{\tau}(\phi(x_g),\xi)\label{eq:KDEs}
\end{alignat}
where $\mathcal{K}_{\tau}(\phi(x),\xi) =  \nicefrac{1}{Z}e^{\frac{\langle \phi(x), \xi \rangle}{\tau}}$ 
is a positive kernel that integrates to $1$ in $\xi$, $\tau>0$ is a temperature parameter that relates to the spread of each kernel, $|S|$ is the cardinality of $S$, and $Z$ is the normalization constant (this becomes irrelevant as it cancels out in the ratios in $\mathcal{L}_D(\phi,G)$ and $\mathcal{L}_G(\phi^\ast,G)$). 
The features $\phi(x)$ are $L^2$-normalized through the projection on the unit hypersphere, \ie, $|\phi(x)|_2=1$. Essentially, we assume that the features are samples of a mixture of von Mises-Fisher distributions, where all concentration parameters are equal to $\nicefrac{1}{\tau}$.

\subsection{Ensuring KDE Accuracy through Data Augmentation}\label{sec:aug}
The KDEs in eq.~\eqref{eq:KDEs} are mixtures of von Mises-Fisher distributions centered around a set of \emph{anchor points}. In the KDE approximation we cannot use the entire dataset $S_r$ as anchor points, because it would be too computationally demanding. Instead, at each iteration of the training procedure we sample a subset (a minibatch) and use this as anchor points.
A fundamental requirement of the KDE approximation is that these sets should be representative of the true distributions $p_r$ or $p_g$. 
However, KDE approximations are in general very poor with high dimensional data, as they require a very large number of anchor points. This is because only the kernels that correspond to anchor points that are ``similar'' to the input sample dominate in the KDE. However, the likelihood of finding these anchor points through uniform sampling becomes extremely small as we grow in data dimensionality.
One way around this problem is to adaptively choose anchor points that provide a good \emph{local} approximation of the pdf at the evaluation points. We create the ``high likelihood'' samples by augmenting the KDE samples, \ie, by transforming samples such that they still belong to their original distribution (so the transformation does not make a real sample unrealistic and a generated example realistic), and they are likely to be ``similar'', but not identical, to the evaluation points. 
For similar reasons, we use a \emph{leave one out} KDE, where we remove the anchor point from the set $S_r$ or $S_g$ that the KDE is being evaluated on. This avoids a bias towards the unlikely case where we sample exactly a point in the anchor point set. We show experimentally that these KDE implementation details are indeed quite important in boosting the effectiveness of the proposed approach.

\subsection{Loss Analysis}\label{sec:loss}
\begin{figure}[t]
  \centering
  \includegraphics[width=.25\textwidth,trim=8cm 3cm 33cm 6cm, clip]{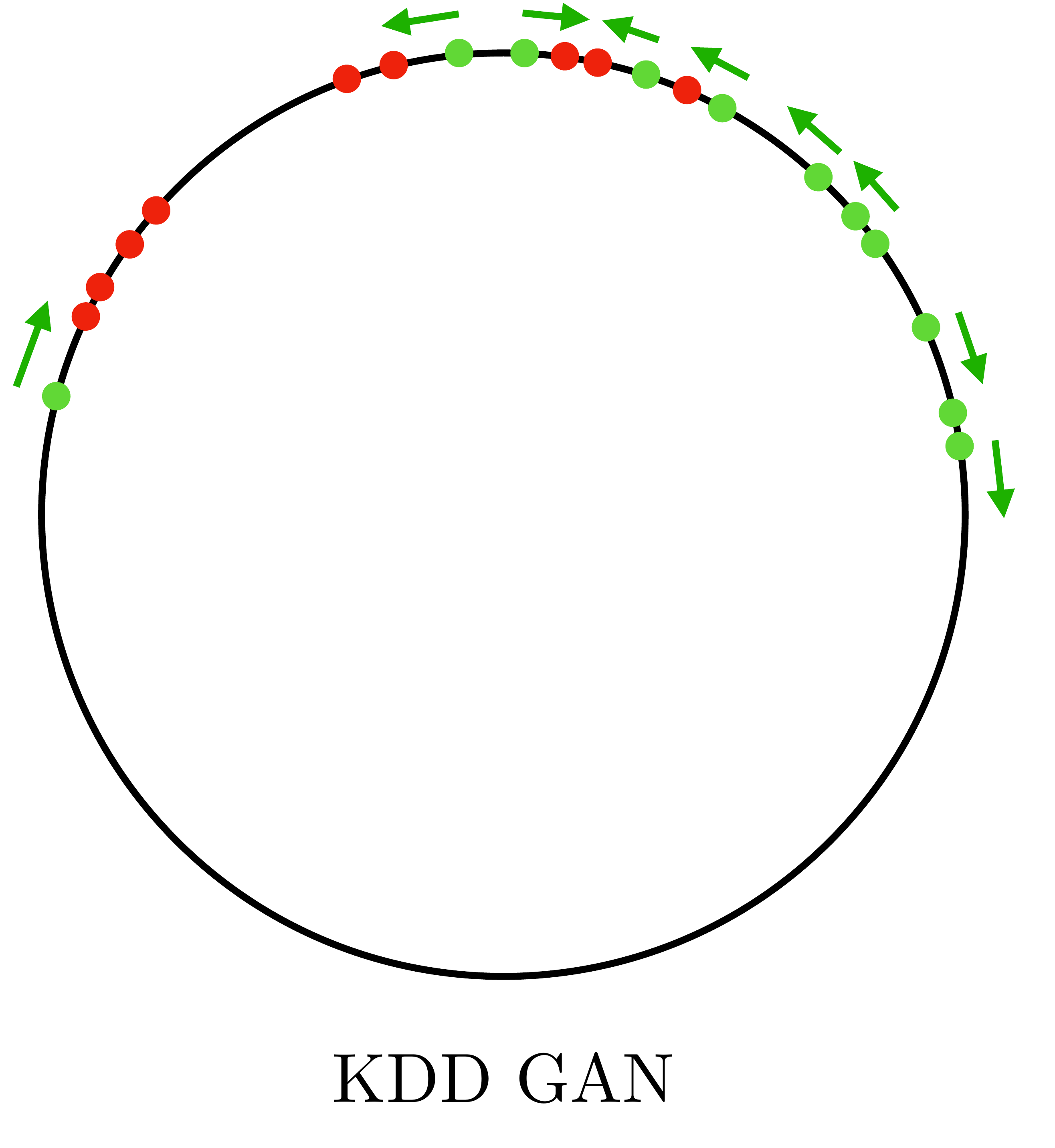}\hspace{1cm}
  \includegraphics[width=.25\textwidth,trim=38cm 3cm 2cm 6cm, clip]{figures/generator_gradient_update.pdf}
  \caption{Comparison of the gradient update directions in feature space between the standard hinge loss and our proposed kernel density loss. Green dots indicate the generated samples and red dots indicate real samples. The green arrows indicate the direction of the update due to the optimization of the generator loss $\mathcal{L}_G$. We can observe that the optimization of the standard hinge loss yields the same direction for all samples (right). In the KDD~GAN the directions change per sample and we can identify two forces: one that pushes generated samples towards the nearest real sample and another that diffuses samples by pushing them away from the other generated samples. \label{fig:gradientUpdate}}
\end{figure}

We also analyze the impact of the proposed loss on the generator training and compare it to the case of the hinge loss standard discriminator of \cite{miyato2018cgans}.
For simplicity, let us consider a discriminator for the standard loss that can be written as the inner product $D_\text{STN}(x) = \phi(x)^\top \theta$, for some $\theta$ vector (this is updated only when we optimize with respect to the discriminator). In the case of the kernel density loss we instead use simply $D_\text{KDE}(x) = \phi(x)$. Suppose that the discriminator is now given and we minimize the loss $\mathcal{L}_G$ with respect to the generator $G$. In the case of a first order optimization method, we obtain the updates for the parameters of the generator through the gradients of $\mathcal{L}_G$, $\frac{\partial \mathcal{L}_G}{\partial G} = \frac{\partial \mathcal{L}_G}{\partial \phi}\frac{\partial \phi}{\partial G}$.
Since in both the standard hinge loss and our loss the term $\frac{\partial \phi}{\partial G}$ is the same, we can reduce the analysis to the study of $\frac{\partial \mathcal{L}_G}{\partial \phi}$. We  obtain: $\frac{\partial \mathcal{L}_G(D_\text{STN})}{\partial \phi} = \theta$ and 
\begin{align}
    \frac{\partial \mathcal{L}_G(D_\text{KDE})}{\partial \phi} = \frac{1}{|S_g|}\sum_{x_g\in S_g}  \frac{\partial \log \hat p_{g}^{D_\text{KDE}}(\phi(x_g))} {\partial \phi} - \frac{ \log \hat p_{r}^{D_\text{KDE}}(\phi(x_g))} {\partial \phi}. \label{eq:KDEupdate}
\end{align}
We can further simplify eq.~\eqref{eq:KDEupdate} to
\begin{align}
    \frac{\partial \mathcal{L}_G(D_\text{KDE})}{\partial \phi} = \frac{1}{|S_g|}\sum_{x_g\in S_g} \frac{\sum_{\bar x_g\in S_g}
    \frac{\partial \mathcal{K}_{\tau}(\phi(\bar x_g),\phi(x_g))}{\partial \phi}}{|S_g|\hat p_{g}^{D_\text{KDE}}(\phi(x_g))} 
    -
    \frac{\sum_{\bar x_r\in S_r} \frac{\partial \mathcal{K}_{\tau}(\phi(\bar x_r),\phi(x_g))}{\partial \phi}}{|S_r|\hat p_{r}^{D_\text{KDE}}(\phi(x_g))}
\end{align}
We observe that the gradient $\frac{\partial \mathcal{L}_G(D_\text{STN})}{\partial \phi}$ 
in the case of the standard loss is simply a constant and this applies to all samples in the minibatch equally. Instead, the gradient \eqref{eq:KDEupdate} changes at each sample in the minibatch (defined by $S_g$). To further simplify the expressions, we can consider that there will be a single dominant term (mode) in the inner sum, consisting of the nearest neighbors in feature space $x_{g}^{\text{NN}}$ and $x_{r}^{\text{NN}}$ to $x_g$, \ie,
\begin{align}\label{eq:forces}
    \frac{\partial \mathcal{L}_G(D_\text{KDE})}{\partial \phi} \simeq \frac{1}{|S_g|}\sum_{x_g\in S_g} \frac{\frac{\partial \mathcal{K}_{\tau}(\phi(x_{g}^{\text{NN}}),\phi(x_g))}{\partial \phi}}{\hat p_{g}^{\phi}(\phi(x_g))} 
    -
    \frac{\frac{\partial \mathcal{K}_{\tau}(\phi( x_{r}^{\text{NN}}),\phi(x_g))}{\partial \phi}}{\hat p_{r}^{\phi}(\phi(x_g))}.
\end{align}
The formula above shows that the gradient update for a sample $x_g$ will depend on two forces, one (term on the left) pushing away from the nearest neighbor(s) in $S_g$ and the other (term on the right) pulling towards the nearest neighbor(s) in $S_r$. More in general, the feature $\phi(x_g)$ will be pulled by the average of all nearby anchor points in $S_r$ and repelled by the average of all nearby anchor points in $S_g$. In particular, the repulsion force is quite important to ensure that generated features spread until they approach real ones.
This is illustrated in Figure~\ref{fig:gradientUpdate}.

\subsection{Class-Conditioning Extension}
We also consider training generative models subject to class-conditioning. Let us denote with $y^{(i)}$ the label corresponding to the real image $x_r^{(i)}$. 
Now, we are interested in the approximation of the quantity $\log \frac{p_r(x,y)}{p_g(x,y)}$, which we can rewrite as $\log \frac{p_r(y|x)p_r(x)}{p_g(y|x)p_g(x)} = \log \frac{p_r(y|x)}{p_g(y|x)} + \log \frac{p_r(x)}{p_g(x)}$. 
The second term is exactly what we used in $\mathcal{L}_D(\phi,G)$ and $\mathcal{L}_G(\phi^\ast,G)$. 
Thus, we can focus on the conditional term $\log \frac{p_r(y|x)}{p_g(y|x)}$. By following \cite{miyato2018cgans}, we assume the linear form $\log \frac{p_r(y|x)}{p_g(y|x)} = y^\top V D(x)$,
where $V$ is a (learned) matrix that defines the embedding for the label $y$.

\subsection{Regularization of the Feature Mapping}
If $\phi$ maps many samples to the same feature, the discrimination task would become less effective. To avoid this scenario, we encourage $\phi_\text{UN}$, the feature mapping before the normalization layer, to be as ``responsive'' as possible to variations around samples of $p_r$ and $p_g$ by introducing the following additional \emph{Jacobian regularization} term
\begin{align}
    \mathcal{L}_\text{Jac} =  \frac{1}{|S_r|}\sum_{x\in S_r \bigcup S_g, \Delta x \sim \mathcal{U}(\mathbb{S}^{d-1})} \left| \delta^{-1} \left|\phi_\text{UN}(x + \delta \Delta x) - \phi_\text{UN}(x) \right|_2 - 1 \right|_1
\end{align}
 where $\delta>0$ is a small scalar and $\Delta x$ is a random unitary direction in image space. $\phi_\text{UN}$ is defined so that $\phi = \phi_\text{UN} / |\phi_\text{UN}|_2$.
This regularization term computes a finite difference approximation of the gradient of $\phi_\text{UN}$ with respect to its input and projects it along the random direction $\Delta x$. 
It preserves as much as possible the volume in feature space, but only for the data on the image distribution. 
In addition, this regularization term prevents the magnification of the output gradient, which is typically associated to a high confidence, and would make the discriminator more susceptible to adversarial inputs. This is a stronger constraint compared the classic gradient penalty \cite{gulrajani2017improved}, since we are implicitly requiring orthonormality for all the rows of the Jacobian, \ie, $\nabla \phi_\text{UN}(x) \nabla \phi_\text{UN}(x)^\top = I_d$.

\subsection{KDD~GAN Formulation} 
Finally, we can put all the terms together and define the generator and discriminator losses via
\begin{equation}
\label{eq:joint-training}
    \mathcal{L}_{G/D} = 
    \gamma \mathcal{L}_{G/D}^\text{KDD} +  \alpha \mathcal{L}_{G/D}^\text{Hinge} +
    \lambda_\nabla \mathcal{L}_\text{Jac},
\end{equation}
where $\gamma $, $\alpha$ and $\lambda_\nabla$ live in $\real^+ \times \{0, 1\} \times \{0, 1e\text{-}5\}$, and where $\text{KDD}$ and $\text{Hinge}$ refer to our KDD loss and the classic hinge loss used in BigGAN for both the generator and discriminator. The training with the lone hinge loss uses $\alpha=1, \gamma =0$; the training with the lone KDD loss uses $\alpha = 0, \gamma=1$; the setting where $\alpha = 1, \gamma>0$ is called \texttt{Joint} training.

\section{Implementation}\label{sec:implementation}

\noindent\textbf{Training Details. }
We evaluate our models on three different datasets: CIFAR10 \cite{krizhevsky2009learning}, Tiny ImageNet and ImageNet $64\times 64$. The Tiny ImageNet \cite{Le2015TinyIN} dataset is a subset of the ILSVRC-2012 ImageNet classification dataset \cite{deng2009imagenet} consisting of $200$ object classes and $500$ training images, $50$ validation images and $50$ test images per class. Unless specified otherwise, we use $\tau = 1$, $\delta = 1e\text{-}3$ and $\lambda_\nabla=1e\text{-}5$. Experiments using data augmentations and the Jacobian regularization are denoted with \textbf{+DA} and \textbf{+JacD} respectively. All training were ran on at most two 2080Ti or one 3090Ti GPUs. Further training details will be available in a code repository.

\noindent\textbf{Architectures. }
The architecture used for our CIFAR10 experiments is the same one\footnote{ \url{https://github.com/ajbrock/BigGAN-PyTorch/}} used in the original BigGAN work by \cite{brock2018large}. For both Tiny ImageNet and ImageNet $64\times 64$, we use the modified SA-GAN \cite{zhang2019self} architecture adopted by \cite{devries2020instance} \footnote{ \url{https://github.com/uoguelph-mlrg/instance\_selection\_for\_gans/}}. We do not use instance selection on CIFAR10 and Tiny ImageNet as we noticed it hurts performance on smaller datasets. For Instance Selection on ImageNet $64\times 64$, we use a retention ratio of 50\%.

\noindent\textbf{Evaluation Metrics. }
Throughout this paper, we evaluate our generative models using Fréchet Inception Distance (FID) \cite{heusel2017gans}, Inception Score (IS) \cite{salimans2016improved}, Density and Coverage \cite{naeem2020reliable}. These metrics are computed using the  original \emph{tensorflow} implementation. As in \cite{devries2020instance} the real moments used for the FID are computed using the whole dataset and not the filtered one. For FID and IS we use 50k generated samples, for Density and Coverage, we use 10k samples for both distributions and $5$ nearest neighbors. Unless specified otherwise, the reported numbers are computed after $100$k iterations for both CIFAR10 and Tiny ImageNet and after 500k iterations for ImageNet $64\times 64$. The batch size used is $64$ for Tiny ImageNet and CIFAR10 and $128$ for ImageNet $64\times 64$. The FID moments are computed on the training set for all datasets. We report the performance of the best model obtained during training.

\noindent\textbf{Differentiable Augmentations. } We use differentiable random brightness, saturation, contrast, translation and cut-out data augmentations proposed by \cite{zhao2020differentiable}. For all our experiments, the loss is computed only on the non-augmented images. The augmented samples are only used for the Kernel Density Estimation. This is an important distinction from the work by \cite{zhao2020differentiable}.

\section{Experiments}\label{sec:exp}

In this section we show the quantitative results obtained on CIFAR10, Tiny ImageNet and ImageNet $64 \times 64$. The best and second best values per metric are highlighted and underlined respectively. Generated samples from one of our best models are shown in Figure \ref{fig:qualitative_examples}. Further qualitative results can be found in the Supplementary Material.
\begin{figure}[t]
    \centering
    \includegraphics[width=\textwidth]{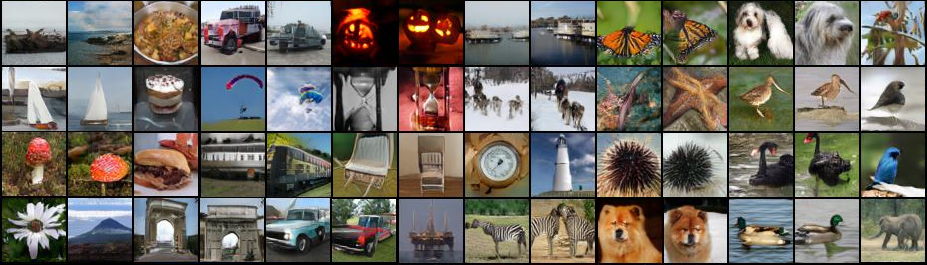}
    \caption{Sample images generated using the Joint\textdagger \ model trained on ImageNet $64\times 64$.}
    \label{fig:qualitative_examples}
\end{figure}

\subsection{Ablation Results}

In Table~\ref{tab:c10_ablation_loss}, we perform various ablations by training BigGAN \cite{brock2018large} on CIFAR10 for 200k iterations each. The three main loss functions used are: the hinge loss \cite{miyato2018cgans}, the KDD loss and the Joint loss.
We study the effects of the parameters associated with the new losses. The first set of experiments studies the effect of the temperature $\tau$ used in the KDD loss. We observe that both high and low values of $\tau$ are problematic. When comparing $\tau=0.05$ to $\tau=5.00$, we observe a trade-off between Image Fidelity (FID) and diversity (IS).
The value of $\tau$ determines the level of blurriness of the KDE.
Additionally, we explore the effect of the Jacobian regularization. We use a coefficient of $\lambda_\nabla = 1e\text{-}5$. Our KDD~GAN using $\tau=1$ with and without the Jacobian regularization outperforms its BigGAN counterpart in both FID and IS. The performance gap is bigger when adding the Jacobian regularization.

The second set of experiments looks at the effect of $\gamma$ during the joint training. We observe that all joint models improve on the baseline in terms of IS. That  improvement correlates positively with $\gamma$ except for $\gamma=10$ where the IS stagnates. The best joint model ($\gamma = 1$) outperforms the baseline also in terms of FID. This highlights the benefit of using the KDD loss as a regularization term. 


Lastly, we train our models without the class-projection head proposed by \cite{miyato2018cgans} and/or without a conditional input for the generator. All models obtained with $\gamma>0$ in the third block in Table~\ref{tab:c10_ablation_loss} outperform the BigGAN baseline in the unconditional setting. This proves that training is not solely driven by the class-projection term in the conditional setting. The difference in performance between unconditional KDD model and the  one with only missing the projection head can be attributed to the slightly higher number of parameters that the latter has since it is still using the class label as input to the generator.

\begin{table}[t]
\caption{Comparison of the various BigGANs trained on CIFAR10.
\textbf{Cond} refers to using the class as input to the generator, while \textbf{Proj} refers to the class-projection loss in ProjGAN \cite{miyato2018cgans}.}
\label{tab:c10_ablation_loss}
\centering
\footnotesize
\begin{tabular}{lccccccccc}
\toprule
\textbf{Experiments} & \textbf{$\tau$} & \textbf{$\gamma$} & \textbf{Cond} & \textbf{Proj} & $\lambda_\nabla$ & \textbf{FID $\downarrow$} & \textbf{IS $\uparrow$} & \textbf{Density $\uparrow$} & \textbf{Coverage $\uparrow$} \\ \midrule
Hinge                & -               & -                 & \checkmark    & \checkmark    & -                & 8.751                     & 8.835                  & 0.966                       & 0.851                        \\
KDD                  & 0.05            & -                 & \checkmark    & \checkmark    & -                & 8.753                     & \textbf{9.233}         & 0.876                       & 0.832                        \\
KDD                  & 1.00            & -                 & \checkmark    & \checkmark    & -                & 8.422                     & \underline{9.155}            & 0.868                       & 0.849                        \\
KDD                  & 5.00            & -                 & \checkmark    & \checkmark    & -                & 8.604                     & 8.852                  & \textbf{0.970}              & 0.862                        \\
KDD + JacD           & 1.00            & -                 & \checkmark    & \checkmark    & 1e-5             & \textbf{7.237}            & 9.029                  & 0.932                       & \underline{0.867}                  \\ \midrule
Joint                & 1.00            & 0.1               & \checkmark    & \checkmark    & -                & 9.144                     & 8.767                  & \underline{0.969}                 & 0.857                        \\
Joint                & 1.00            & 0.5               & \checkmark    & \checkmark    & -                & 8.795                     & 8.920                  & 0.922                       & 0.855                        \\
Joint                & 1.00            & 1.0               & \checkmark    & \checkmark    & -                & \underline{7.932}               & 9.046                  & 0.968                       & \textbf{0.868}               \\
Joint                & 1.00            & 10.0              & \checkmark    & \checkmark    & -                & 8.352                     & 9.102                  & 0.930                       & 0.857                        \\ \midrule
KDD                  & 0.05            & -                 & \checkmark    & -             & -                & 13.668                    & 8.274                  & 0.722                       & 0.711                        \\
Hinge                & -               & -                 & -             & -             & -                & 17.782                    & 8.120                  & 0.692                       & 0.686                        \\
KDD                  & 0.05            & -                 & -             & -             & -                & 15.828                    & 8.326                  & 0.620                       & 0.650                        \\
Joint                & 0.05            & 1.0               & -             & -             & -                & 14.394                    & 8.532                  & 0.662                       & 0.712                        \\ \bottomrule
\end{tabular}
\end{table}

\subsection{Generative Learning on CIFAR10  (Class-Conditional)}


In Table~\ref{tab:C10-eval-100k}, we compare the performance of different variations of our KDD~GAN with a BigGAN baseline and the numbers reported by \cite{kang2020contragan} for a selection of their best models.
The KDD~GAN outperforms the BigGAN baseline for both IS and FID. Also it drastically improves its FID when using augmentations as described in section~\ref{sec:aug}. $\text{Augmentation} \times N$ means that an additional $N \times batch size$ augmented images are used for the KDE anchor points. We observe that on CIFAR10, the amount of augmentations correlates positively with a significant improvement of the FID. In the case of the Jacobian regularization the results are mixed. It seems to improve the performance of the KDD model, but it also negatively impacts performance when used in combination with data augmentation. The Jacobian regularization may be too strict a requirement, as the dimension $K$ of the gradient of $\phi$ is smaller than the dimension $d$ of the images, and perhaps a more flexible loss term could work better. 


\begin{table}[t]
\centering
\caption{Experimental results on CIFAR10. The values shown below are obtained after 100k iterations. We show the benefit of adding various augmentation factors for the KDD setting. We also explore the effect of the Jacobian regularization. $^\star$ are numbers reported by \cite{kang2020contragan}.}
\footnotesize
\label{tab:C10-eval-100k}
\begin{tabular}{lcccccc}
\toprule
\textbf{Experiments} & $\lambda_\nabla$ & \textbf{DA} & \textbf{FID $\downarrow$} & \textbf{IS $\uparrow$}& \textbf{Density $\uparrow$} & \textbf{Coverage $\uparrow$} \\ \midrule
ContraGAN$^\star$                         & -                     & -                    & 8.065                 & 9.729                & -                         & -                          \\
ContraGAN + DiffAugment$^\star$           & -                     & -                    & 7.193                 & \underline{9.996}          & -                         & -                          \\
BigGAN + DiffAugment$^\star$              & -                     & -                    & \textbf{7.157}        & 9.775                & -                         & -                          \\
BigGAN + CR$^\star$                       & -                     & -                    & \underline{7.178}           & \textbf{10.380}      & -                         & -                          \\ \midrule
Hinge loss                                 & -                         & -           & 8.859          & 8.814          & 0.917            & 0.841             \\
KDD                                      & -                         & -           & 8.375          & 8.901          & 0.875            & 0.845             \\
KDD + DA                       & -                         & \checkmark  & 7.089          & 9.250          & 0.893            & 0.860             \\
KDD + DA $\times 3$            &                           & $\times 3$  & \underline{6.063}    & 9.280          & \underline{0.951}      & \underline{0.892}       \\
KDD + DA $\times 7$            & -                         & $\times 7$  & \textbf{5.713} & \textbf{9.389} & \textbf{0.968}   & \textbf{0.899}    \\ 
KDD + JacD                               & 1e-5                      & -           & 7.944          & 8.959          & 0.895            & 0.847             \\
KDD + JacD + DA $\times 7$     & 1e-5                      & $\times 7$  & 6.713          & \underline{9.333}    & 0.9000           & 0.875             \\ \bottomrule
\end{tabular}
\end{table}


\subsection{Generative Learning on ImageNet (Class-Conditional)}

\noindent\textbf{Tiny ImageNet. }
Table~\ref{tab:tinyIN-eval} shows the performance of our models on Tiny ImageNet compared to the SA-GAN baseline and the best models reported by \cite{kang2020contragan}.
The KDD~GAN outperforms the baseline for all settings. On one hand, similarly to CIFAR10, using additional  augmented images for the KDE results in a significant boost in performance. Indeed the KDD~GAN with DA $\times 3$ outperforms ContraGAN in terms of FID and IS. On the other hand, the additional Jacobian regularization is not helpful. The only exception being the joint training ($\gamma=0.5$) without data augmentation and the joint training with $\gamma = 1$ and data augmentation where the Jacobian regularization introduces a slight performance boost.  

\begin{table}[t]
\centering
\caption{Results on Tiny ImageNet. We compare the baseline to both the KDD and joint trainings. We also explore the effect of adding the Jacobian regularization on D and show the effect of using more augmentations for the density estimation. $^\star$ are numbers reported by \cite{kang2020contragan}.}
\label{tab:tinyIN-eval}
\footnotesize
\begin{tabular}{lccccccc}
\toprule
\textbf{Experiments} & \textbf{$\gamma$} & $\lambda_\nabla$ & \textbf{DA} & \textbf{FID $\downarrow$} & \textbf{IS $\uparrow$} & \textbf{Density $\uparrow$} & \textbf{Coverage $\uparrow$} \\ \midrule
ContraGAN$^\star$                        & -                          & -                      & -                    & 27.027                & 13.494               & -                         & -                 \\
ContraGAN  + DiffAugment$^\star$           & -                          & -                      & -                    & \textbf{15.755}                & \textbf{17.303}               & -                         & -                 \\ \midrule
Hinge loss                       & -                          & -                         & -                    & 29.525                             & 11.048                          & 0.520                                & 0.516                        \\
KDD                              & -                          & -                         & -                    & 24.022                             & 13.204                          & 0.658                                & 0.613                        \\
KDD + DA                         & -                          & -                         & \checkmark           & \underline{ 20.204}                       & \underline{ 14.100}                    & \underline{ 0.673}                          & \underline{ 0.663}                  \\
KDD + DA $\times 3$              & -                          & -                         & $\times 3$           & \textbf{18.261}                    & \textbf{14.943}                 & \textbf{0.716}                       & \textbf{0.683}               \\
KDD + JacD                       & -                          & 1e-5                      & -                    & 25.504                             & 13.215                          & 0.597                                & 0.595                        \\
KDD + JacD + DA                  & -                          & 1e-5                      & \checkmark           & 20.717                             & 13.787                          & 0.630                                & 0.645                        \\
Joint + DA                       & 1                          &                           & \checkmark           & 22.854                             & 13.421                          & 0.591                                & 0.613                        \\
Joint + JacD + DA                & 1                          & 1e-5                      & \checkmark           & 21.512                             & 13.728                          & 0.639                                & 0.627                        \\
Joint                            & 0.5                        & -                         & -                    & 24.341                             & 13.337                          & 0.626                                & 0.614                        \\
Joint + JacD                     & 0.5                        & 1e-5                      & -                    & 23.854                             & 13.251                          & 0.651                                & 0.617                        \\
\bottomrule
\end{tabular}
\end{table}

\noindent\textbf{ImageNet $\mathbf{64 \times 64}$. }
Table~\ref{tab:I64-eval} shows our experimental results on ImageNet $64\times 64$. We compare our models to the SA-GAN baseline and the numbers reported by \cite{devries2020instance}. For all our trained models, we use Instance Selection \cite{devries2020instance} with a retention ratio of 50\%.

We observe that the baseline outperforms our KDD~GAN even with additional augmentations and regularization. It also note-worthy that in this setting, although a small amount of data augmentation seems to help, adding more is not necessarily beneficial. The high  level of diversity in ImageNet both in terms of number of classes and samples might be limiting the effectiveness of our density estimation given the relatively small batch size used.

Nevertheless, all joint training models outperform the hinge-based models in terms of IS and most outperform our SA-GAN baseline in terms of FID. Interestingly, the best model is the Joint\textdagger \   model where $\hat{p}_r$ is estimated using features computed during the last discriminator step. This suggests that using a memory bank for the features might be a promising extension of this work.

\begin{table}[t]
\centering
\caption{Quantitative results on ImageNet $64\times 64$. We explore the use of augmentation, Jacobian regularization and Joint training. \textdagger \  refers to a setting where the feature $\phi(x_r)$ were computed using the weights from the previous discriminator update step. $^\star$ are numbers reported by \cite{devries2020instance}. }
\label{tab:I64-eval}
\footnotesize
\begin{tabular}{lcccccc}
\toprule
\textbf{Experiments} & \textbf{$\gamma$} & \textbf{DA} & \textbf{FID $\downarrow$} & \textbf{IS $\uparrow$} & \textbf{Density $\uparrow$} & \textbf{Coverage $\uparrow$} \\ \midrule
SA-GAN + IS @ 50 \%$^\star$     & -                 & -           & \underline{9.63}                     & 31.04                  & \underline{1.07}                  & 0.88                         \\
FQ-BigGAN$^\star$               & -                 & -           & \underline{9.67}                     & 25.96                  & -                           & -                            \\ \midrule
Hinge loss                        & -                 & -           & 10.452                    & 32.869                 & 1.034                       & 0.877                        \\
KDD                             & -                 & -           & 12.570                    & 31.404                 & 0.953                       & 0.850                        \\
KDD + DA                        & -                 & \checkmark  & 12.367                    & 31.069                 & 0.954                       & 0.861                        \\
KDD + DA x3                     & -                 & $\times 3$  & 14.680                    & 27.949                 & 0.928                       & 0.810                        \\
KDD + JacD                      & -                 & -           & 12.651                    & 31.188                 & 0.938                       & 0.850                        \\
KDD + JacD + DA                 & -                 & \checkmark  & 79.790                    & 10.603                 & 0.376                       & 0.139                        \\ \midrule
Joint                           & 0.5               & -           & 10.544                    & 33.447                 & 1.017                       & 0.879                        \\
Joint \textdagger               & 0.5               & -           & \textbf{9.450}            & \textbf{35.648}        & \underline{1.070}                 & 0.897                        \\
Joint + DA                      & 0.5               & \checkmark  & 10.111                    & 33.494                 & 1.048                       & \underline{0.891}                  \\
Joint + JacD                    & 0.5               & -           & 10.242                    & \underline{35.120}           & \textbf{1.072}              & \underline{0.891}                  \\
Joint + JacD + DA           & 1.0            & \checkmark  & 9.702                     & 34.619                 & 1.062                       & 0.892                        \\ \bottomrule
\end{tabular}
\end{table}


\section{Limitations and Future Work}\label{sec:limit}

One of the major challenges in the use of KDD~GAN is to ensure that the anchor points for the KDE are representative for the evaluation points. In our experiments between Tiny ImageNet and  ImageNet $64\times 64$, we observe that the performance of KDD~GAN is sensitive to the anchor points set size, the number of augmentations, and the complexity of the dataset seems to play a role too. 
Also, with large datasets the impact of samples at the tails of the distribution on the KDE approximation is unclear. More in general, it might be necessary to design sampling strategies for the samples used for the KDE estimation: Some options are using a memory bank or  sampling using k-NN. Another direction to evaluate is the role of the class projection in the training. We chose to separate the category aspect from the unlabeled problem not only because it would make KDD~GAN suitable for unsupervised learning, but also because it would not require large minibatches as the current KDE ignores completely the class labels. It would be interesting to evaluate the performance in the case where the loss with class labels is entirely based on the KDE.
Finally, as mentioned in the introduction, KDD~GAN can be combined with other techniques and regularization methods that are known to improve the performance of the GAN training, such as Consistency Regularization of \cite{zhang2019consistency} and Differentiable Augmentation of \cite{zhao2020differentiable}. 
We leave these investigations to future work. 



\section{Related work}
\paragraph{Generative Modeling.} The goal in generative learning is to model the data distribution $p(x)$ or $p(x,y)$. This can be achieved either by learning an explicit representation of $p(x)$ or by learning a sampling protocol from said distribution. The former approach is based on maximizing the likelihood of the data samples within a family of known distribution $\mathcal{P}$. One prominent line of research that falls under this umbrella is flow-based modeling \cite{kobyzev2020normalizing, dinh2016density}. In this setting a known distribution, typically a multi-variate Gaussian, is mapped to the data space through a parametric invertible function, which  defines a push-forward distribution using the change of variables formula. In the latter approach, one of the most prominent models  is Generative Adversarial Networks \cite{goodfellow2014generative}, where one uses a generator network to map samples from a known distribution, typically a multi-variate Gaussian, to samples from the data distribution. The training relies on an adversarial min-max game, where a discriminator is trained to distinguish real samples from fake ones while the generator is trained to fool it. The link to the data distribution is implicitly defined through the saddle points of the min-max optimization, where the generator loss corresponds to a known statistical divergence, \eg, the Jensen-Shannon Divergence (JSD) \cite{goodfellow2014generative}.\\
\noindent\textbf{Improving GANs.} Ever since their inception, GANs have gone through various iterations and improvements. Works such as \cite{arjovsky2017wasserstein} and \cite{nowozin2016f} focus on training the generator to minimize other statistical divergences that exhibit better properties compared to the JSD in the original work. One complementary line of research explores additional regularization terms such as using a gradient penalty for the discriminator \cite{mescheder2018training}, consistency regularization \cite{zhang2019consistency} or  differentiable augmentations \cite{zhao2020differentiable} to various degrees of success. A recent addition to this list are methods that capture more structure into the latent representation of the discriminator through the use of Contrastive Learning \cite{kang2020contragan, jeong2021contrad, yu2021dual}. One such example is ContraGAN \cite{kang2020contragan}, where the authors introduce a new regularization term, called 2C loss, based on the NT-Xent loss \cite{chen2020simple} used commonly in Contrastive Learning. The introduced loss term aims at capturing the data-to-data and data-to-class relations in the dataset.\\
\noindent\textbf{Contrastive Learning.} Contrastive Learning methods aim at maximizing the similarity between pairs of positive samples, \eg, different augmentations of the same image or samples taken from the same class, while minimizing the similarity between pairs of negative samples, \eg, augmentations of different images or samples taken from different classes with applications ranging from metric learning to self-supervised training \cite{le2020contrastive}. One of the most commonly used losses in this context is the Normalized Temperature-scaled Cross Entropy Loss (NT-Xent) \cite{chen2020simple}. \cite{wang2020understanding} shows this loss can be split into two terms: an alignment and a uniformity constraints. The former encourages the alignment of positive pairs of samples whereas the latter encourages the feature representation to be uniformly distributed on the unit-sphere. The uniformity constraint stems from maximizing the entropy of the distribution of the features, which is computed through kernel density estimation using the von Mises-Fisher kernel on the unity sphere. This interpretation of the uniformity constraint is validated experimentally by replacing the entropy term with other statistical divergences between the features distribution and a uniform one.\\
\noindent\textbf{Instance Selection for GANs.} Training GANs on large scale datasets is a challenging task. State of the art models such as BigGAN \cite{brock2018large} require a substantial amount of compute resources. Moreover, many of them require a post-hoc processing to reduce spurious samples. \cite{devries2020instance} proposes to tackle both issues by filtering the dataset using instance selection. They argue that the model's capacity is wasted on low density regions of the empirical distributions of the data. Their results show that instance selection allows to train better GAN models using substantially fewer parameters and training time.

\section{Conclusions}\label{sec:conclusion}
We have introduced KDD~GAN, a novel GAN formulation based on a KDE loss defined in feature space. The proposed method is flexible as it can be integrated with other techniques and keeps the computational complexity and memory requirements at bay. 
We also show experimentally that KDD~GAN yields a substantial improvement in FID and IS under several settings over the same baseline method (BigGAN/SA-GAN).
Because our formulation explicitly models the data distribution, it can also directly control the biases learnt from the dataset. Thus, our framework could be used to potentially influence either in a negative or positive way the training and its consequent societal impact.
This remains to be further explored.

\bibliography{biblio}
\FloatBarrier
\clearpage

\maketitle

\appendix

\section{Empirical Analysis of the KDD Loss}
\begin{figure}[ht]
    \centering
    \subfigure[Initial state \label{fig:kdd-init}]{\includegraphics[width=0.32\textwidth]{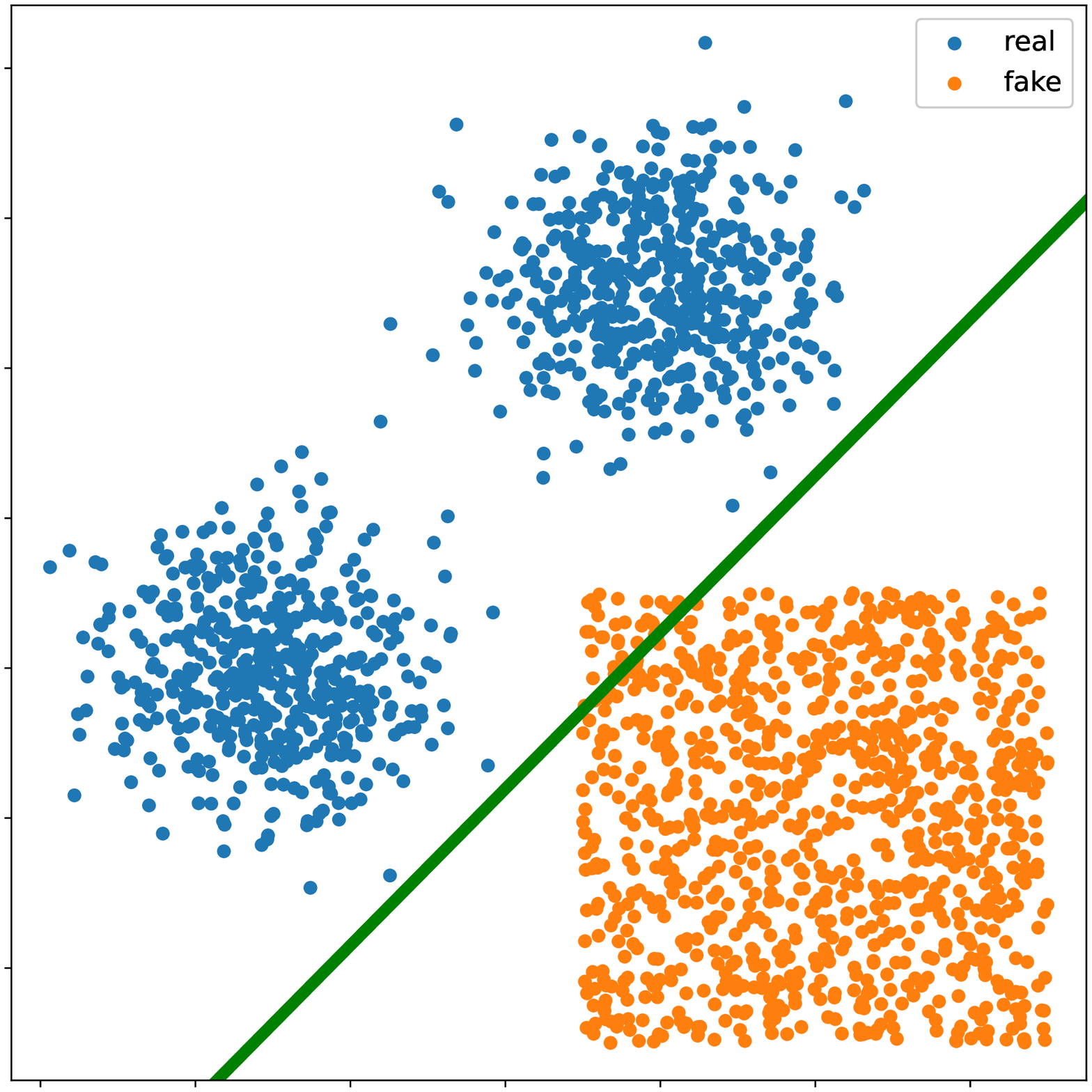}}
    \subfigure[Hinge loss \label{fig:kdd-clf}]{\includegraphics[width=0.32\textwidth]{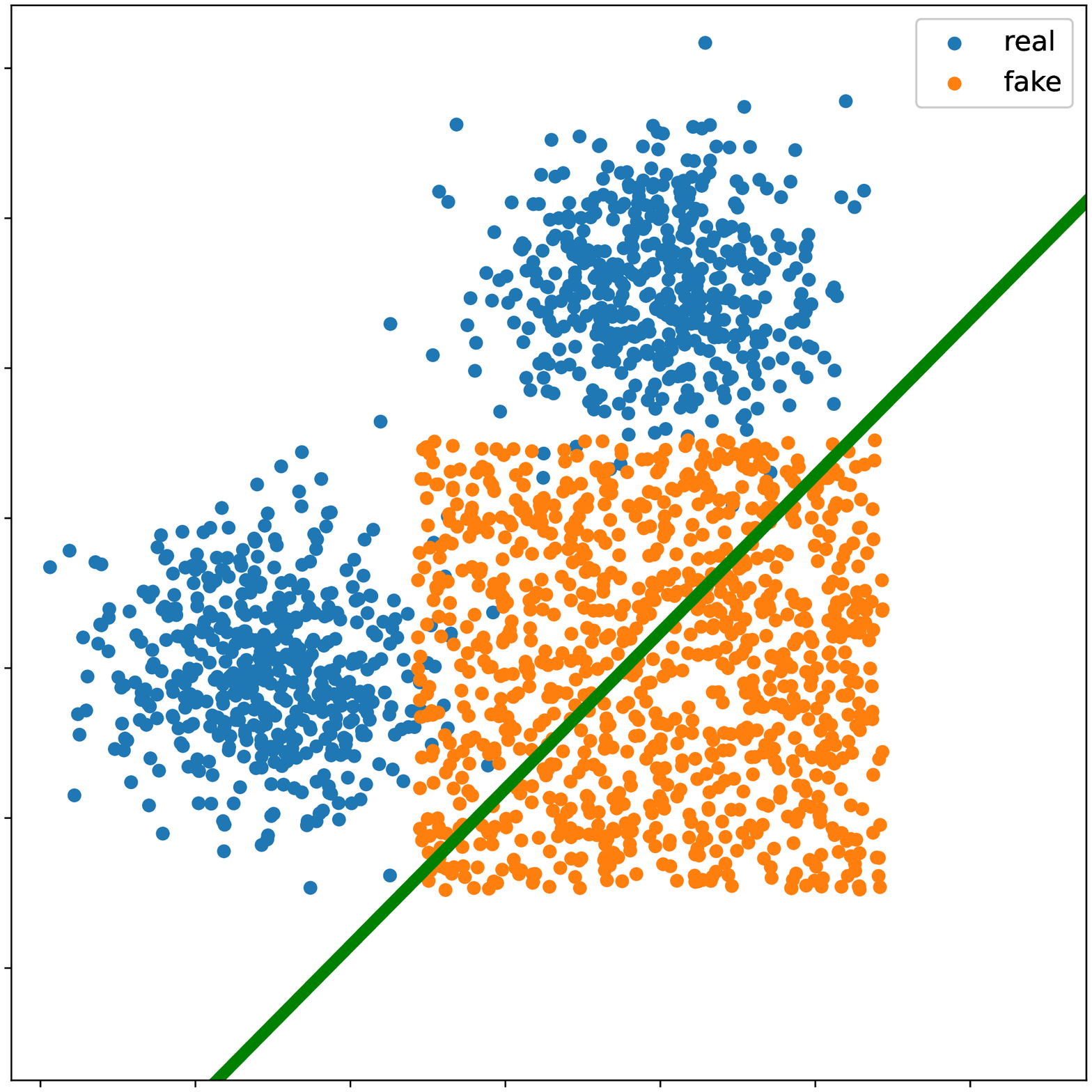}}
    \subfigure[KDD loss \label{fig:kdd-kdd}]{\includegraphics[width=0.32\textwidth]{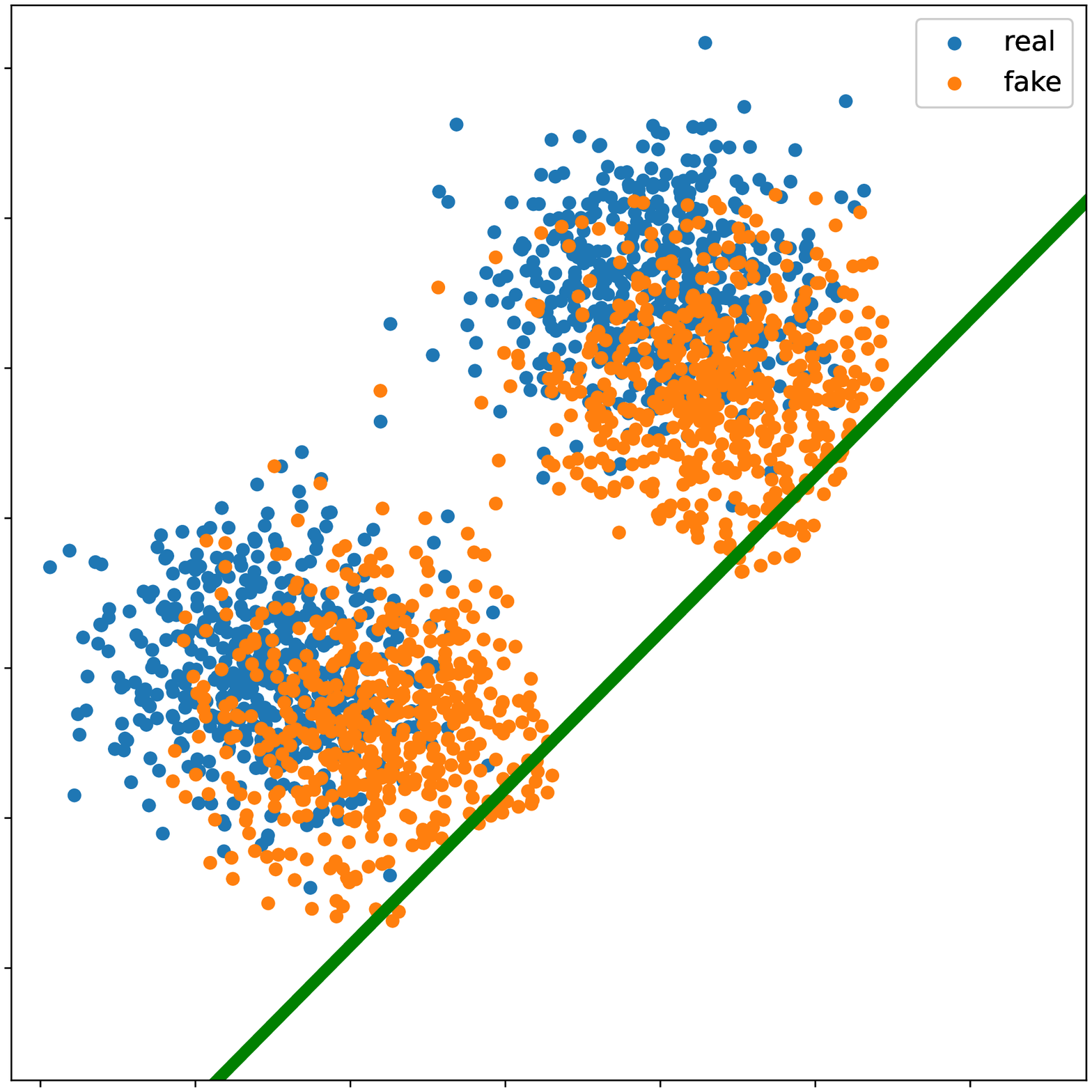}}
    \caption{\textbf{Illustration of the difference between the Hinge loss and KDD loss during the Generator update.} The blue and orange point clouds represent the real and fake samples. The initial positions of the samples are shown in Figure~\ref{fig:kdd-init}. The green line in all three sub-figures represents the decision boundary associated with the optimal linear classifier separating the two distributions in the initial state. Figures~\ref{fig:kdd-clf} and ~\ref{fig:kdd-kdd} show the updated positions of the fake samples using the Hinge loss and KDD loss respectively.}
    \label{fig:illustration-kdd}
\end{figure}

In Figure~\ref{fig:illustration-kdd}, we illustrate the difference between the Hinge and KDD losses already described in Section~\ref{sec:loss}. We consider two point clouds in 2D representing the real and fake push-forward distributions. In this examples, the real point cloud is designed to have two Gaussian modes, while the fake one starts off with one uniformly sampled square mode. We first find the optimal linear classifier separating the two point clouds through gradient descent. The corresponding decision boundary is represented by the green line in Figure~\ref{fig:illustration-kdd}. We then optimize the features of the fake samples with respect to the Hinge loss and the KDD loss. In this example, since we are focusing on the optimization of the Generator, it is not important whether the feature mapping is normalized or not.\footnote{The purpose of the normalization is to prevent the Discriminator from converging to degenerate solutions, where the space collapses.} Thus, for visualization purposes, we work with 2D features. Also, we forgo the feature normalization and use a Gaussian kernel with $\sigma = 1$ for the KDE, \ie, $\mathcal{K}(\phi,\xi) \propto \exp{-\nicefrac{|\phi-\xi|^2}{2}}$.  

The minimization of the Hinge loss simply results in translating the fake point cloud without changing its internal structure as shown in Figure~\ref{fig:kdd-clf}. In contrast, the KDD loss encourages the fake samples to head towards the closest real mode as shown in Figure~\ref{fig:kdd-kdd}. For both of the losses, the optimization was ran using SGD~\cite{bottou1998online}  for 200 iterations with a learning rate of 10. 1000 samples were used for both the real and fake point clouds. Note that for a frozen Discriminator, updating the Generator using the Hinge loss can result in overshooting the real point cloud, since the translation vector is constant for all subsequent Generator updates. In fact, the optimum is to translate the fake point cloud to infinity. This makes the Generator update with respect to the Hinge loss less well-behaved than its KDD counterpart since the latter does not introduce such instability.

\section{Quantitative Results}
We report the performance numbers for a few additional configurations on Tiny ImageNet and ImageNet $64 \times 64$ in Tables \ref{tab:I64-extra} and \ref{tab:tinyIN-extra} respectively. In both tables, KDD, Joint, JacD and DA refer to  training using the KDD loss ($\gamma = 1$ \& $\alpha = 0$), the joint training ($\alpha = 1$), the Jacobian regularization and the use of data augmentation for Kernel Density Estimation respectively.

As stated in the main paper, we observe in Table~\ref{tab:I64-extra} that using data augmentation for the kernel density estimation always improves performance with respect to the FID. However, increasing the amount of augmentations does not yield the same boost in performance on ImageNet $64 \times 64$ as on other datasets. Additionally, while the Jacobian Regularization seems harmful when used in the KDD setting, it consistently improves performance when used during the Joint training. 

For the Tiny ImageNet dataset, the results reported in Table~\ref{tab:tinyIN-extra} show that data augmentation is almost always beneficial. Notably, increasing the amount of augmentations yields a signification boost in performance.  On the other hand, the results for the Jacobian regularization are mixed, the best case scenario being \texttt{Joint + DA} compared to \texttt{Joint + JacD + DA}.


\begin{table}[t]
\centering
\caption{Additional results on Image $64 \times 64$. \textdagger \  refers to a setting where the feature $\phi(x_r)$ were computed using the weights from the previous discriminator update step.  $^\star$ are numbers reported by \cite{devries2020instance}. }
\label{tab:I64-extra}
\begin{tabular}{lccccccc}
\toprule
\textbf{Experiments}                 & \textbf{$\gamma$} & $\lambda_\nabla$ & \textbf{DA} & \textbf{FID $\downarrow$} & \textbf{IS $\uparrow$} & \textbf{Density $\uparrow$} & \textbf{Coverage $\uparrow$} \\ \midrule
SA-GAN + IS @ 50 \%$^\star$ & -                 & -               & -           & \underline{9.63}                      & 31.04                  & \underline{1.07}                        & \underline{0.88}                         \\
FQ-BigGAN$^\star$           & -                 & -               & -           & \underline{9.67}                      & 25.96                  & -                           & -                            \\ \midrule
Hinge loss                    & -                 & -               & -           & 10.452                    & 32.869                 & 1.034                       & 0.877                        \\ \midrule
KDD                         & -                 & -               & -           & 12.570                    & 31.404                 & 0.953                       & 0.850                        \\
KDD + DA                    & -                 & -               & \checkmark  & 12.367                    & 31.069                 & 0.954                       & 0.861                        \\
KDD + DA                    & -                 & -               & $\times 3$  & 14.680                    & 27.949                 & 0.928                       & 0.810                        \\
KDD + JacD                  & -                 & 1e-5            & -           & 12.651                    & 31.188                 & 0.938                       & 0.850                        \\
KDD + JacD + DA             & -                 & 1e-5            & \checkmark  & 79.790                    & 10.603                 & 0.376                       & 0.139                        \\ \midrule
Joint                       & 1                 & -               & -           & 11.387                    & 32.471                 & 0.991                       & 0.872                        \\
Joint + DA                   & 1                 & -               & \checkmark  & 10.385                    & 33.753                 & 1.048                       & 0.880                        \\
Joint + JacD                & 1                 & 1e-5            & -           & 10.320                    & 34.296                 & 1.010                       & 0.868                        \\
Joint + JacD + DA           & 1                 & 1e-5            & \checkmark  & 9.702                     & 34.619                 & 1.062                       & \underline{0.892}                        \\ \midrule
Joint                       & 0.5               & -               & -           & 10.544                    & 33.447                 & 1.017                       & 0.879                        \\
Joint \textdagger           & 0.5               & -               & -           & \textbf{9.450}            & \textbf{35.648}        & \underline{1.070}                       & \textbf{0.897}                        \\
Joint + DA                  & 0.5               & -               & \checkmark  & 10.111                    & 33.494                 & 1.048                       & \underline{0.891}                        \\
Joint + JacD                & 0.5               & 1e-5            & -           & 10.242                    & \underline{35.120}                & \textbf{1.072}                       & \underline{0.891}                        \\
Joint + JacD + DA           & 0.5               & 1e-5            & \checkmark  & 10.010                    & 34.074                 & 1.053                       & 0.889                        \\ \bottomrule
\end{tabular}
\end{table}

\begin{table}[h]
\centering
\caption{Additional Results on Tiny ImageNet. $^\star$ are numbers reported by \cite{kang2020contragan}.}
\label{tab:tinyIN-extra}
\begin{tabular}{lccccccc}
\toprule
\textbf{Experiments}             & \textbf{\textbf{$\gamma$}} & \textbf{$\lambda_\nabla$} & \textbf{\textbf{DA}} & \textbf{\textbf{FID $\downarrow$}} & \textbf{\textbf{IS $\uparrow$}} & \textbf{\textbf{Density $\uparrow$}} & \textbf{Coverage $\uparrow$} \\ \midrule
ContraGAN$^\star$                & -                          & -                        & -                    & 27.027                             & 13.494                          & -                                    & -                            \\
ContraGAN  + DiffAugment$^\star$ & -                          & -                        & -                    & \textbf{15.755}                    & \textbf{17.303}                 & -                                    & -                            \\ \midrule
Hinge loss                         & -                          & -                        & -                    & 29.525                             & 11.048                          & 0.520                                & 0.516                        \\
KDD                              & -                          & -                        & -                    & 24.022                             & 13.204                          & 0.658                                & 0.613                        \\
KDD + DA                         & -                          & -                        & \checkmark           & \underline{20.204}                       & \underline{14.100}                    & \underline{0.673}                          & \underline{0.663}                  \\
KDD + DA $\times 3$              & -                          & -                        & $\times 3$           & \textbf{18.261}                    & \textbf{14.943}                 & \textbf{0.716}                       & \textbf{0.683}               \\
KDD + JacD                       & -                          & 1e-5                     & -                    & 25.504                             & 13.215                          & 0.597                                & 0.595                        \\
KDD + JacD + DA                  & -                          & 1e-5                     & \checkmark           & 20.717                             & 13.787                          & 0.630                                & 0.645                        \\ \midrule
Joint                            & 1                          & -                        & -                    & 25.709                             & 13.124                          & 0.595                                & 0.582                        \\
Joint + DA                       & 1                          &                          & \checkmark           & 22.854                             & 13.421                          & 0.591                                & 0.613                        \\
Joint + JacD                     & 1                          & 1e-5                     & -                    & 26.369                             & 13.169                          & 0.582                                & 0.582                        \\
Joint + JacD + DA                & 1                          & 1e-5                     & \checkmark           & 21.512                             & 13.728                          & 0.639                                & 0.627                        \\ \midrule
Joint                            & 0.5                        & -                        & -                    & 24.341                             & 13.337                          & 0.626                                & 0.614                        \\
Joint + DA                       & 0.5                        &                          & \checkmark           & 23.357                             & 12.918                          & 0.619                                & 0.621                        \\
Joint + JacD                     & 0.5                        & 1e-5                     & -                    & 23.854                             & 13.251                          & 0.651                                & 0.617                        \\
Joint + JacD + DA                & 0.5                        & 1e-5                     & \checkmark           & 23.928                             & 13.059                          & 0.575                                & 0.594                        \\ \bottomrule
\end{tabular}
\end{table}

\FloatBarrier
\section{Examples of Generated Images}
In this section, we show non-cherry picked images generated by our Hinge loss baseline and our best model per dataset. The truncation trick was not used \cite{brock2018large}. In all figures, each row represents a different class starting with the first class in the top row down to the last class in the bottom row.

\subsection{CIFAR10}


\begin{figure}[h]
    \centering
    \subfigure[Hinge loss \label{fig:C10Baseline}]{\includegraphics[width=0.49\textwidth]{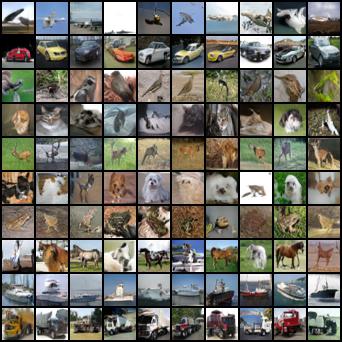}}
    \subfigure[KDD + Aug $\times 7$ \label{fig:C10Top}]{\includegraphics[width=0.49\textwidth]{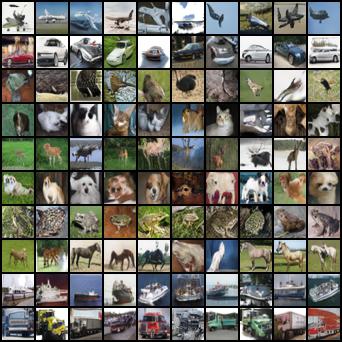}}
    \caption{Samples generated using the Hinge loss model and the KDD + Aug $\times 7$ model trained on CIFAR10 (one class per row).}
    \label{fig:my_label}
\end{figure}

\FloatBarrier
\clearpage
\subsection{Tiny ImageNet}
\begin{figure}[ht]
    \centering
    \includegraphics[height=0.8\textheight]{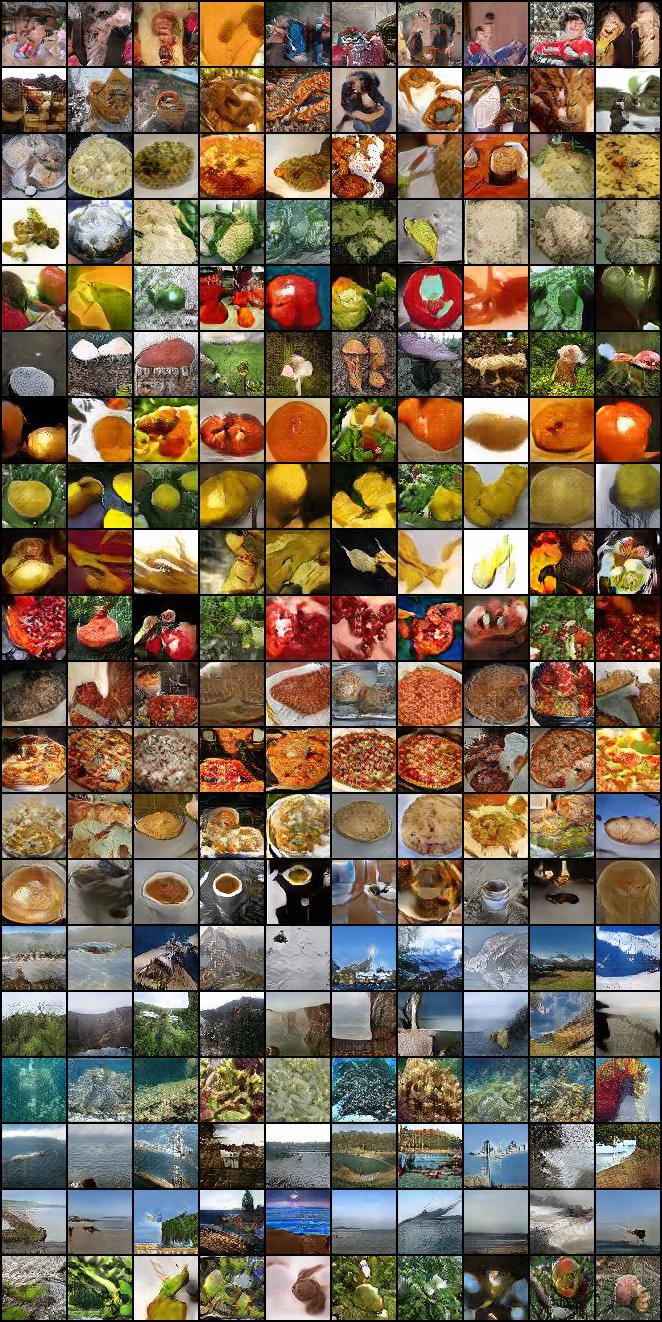}
    \caption{Samples generated using the Hinge loss model trained on Tiny ImageNet  for the classes 181-200 (one class per row).}
    \label{fig:tinyINBaseline}
\end{figure}

\begin{figure}[ht]
    \centering
    \includegraphics[height=0.8\textheight]{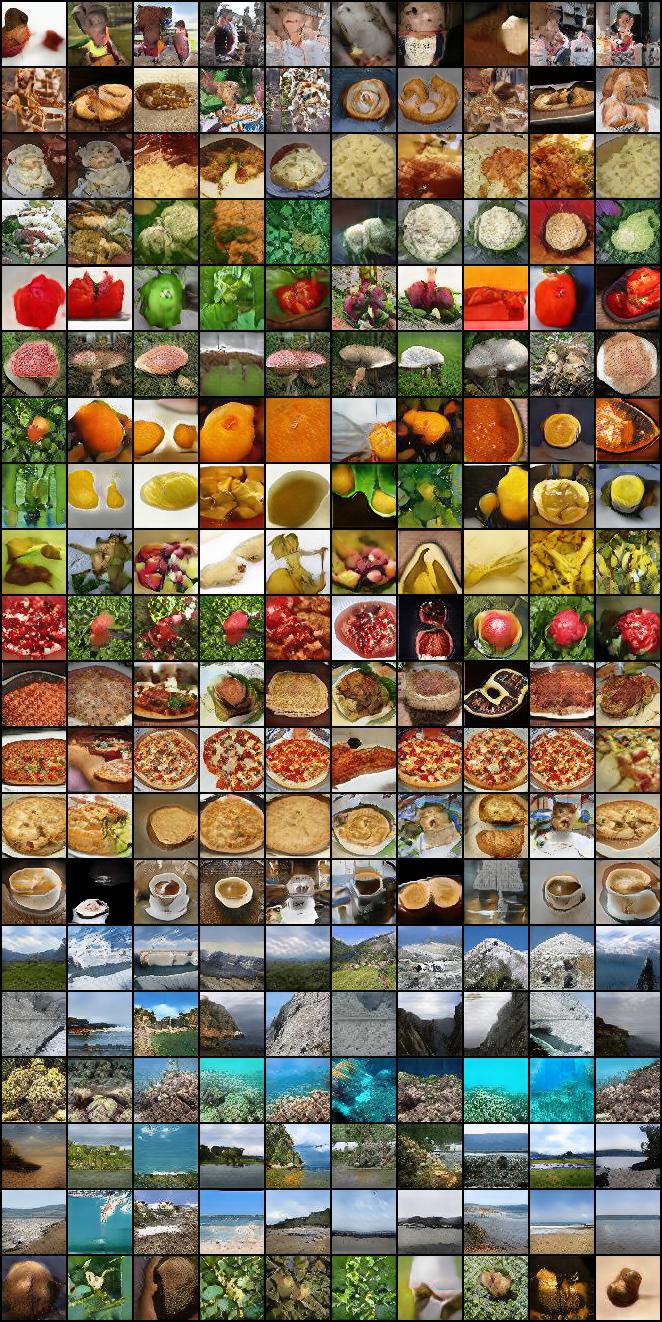}
    \caption{Samples generated using the KDD+Aug $\times 3$ model trained on Tiny ImageNet for the classes 181-200 (one class per row).}
    \label{fig:tinyINTop}
\end{figure}

\clearpage
\subsection{ImageNet 64x64}

\begin{figure}[ht]
    \centering
    \includegraphics[height=0.8\textheight]{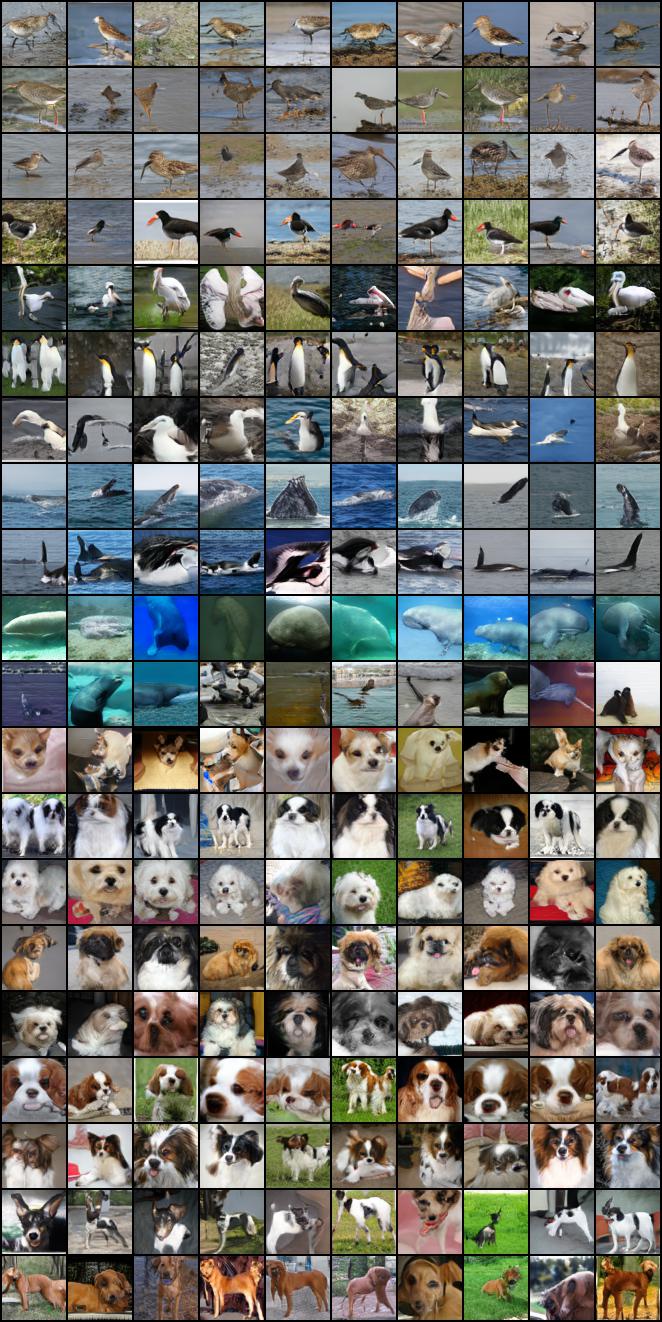}
    \caption{Samples generated using Hinge loss model trained on ImageNet $ 64 \times 64$ for the classes 141-160 (one class per row).}
    \label{fig:I64Baseline2}
\end{figure}
\begin{figure}[ht]
    \centering
    \includegraphics[height=0.8\textheight]{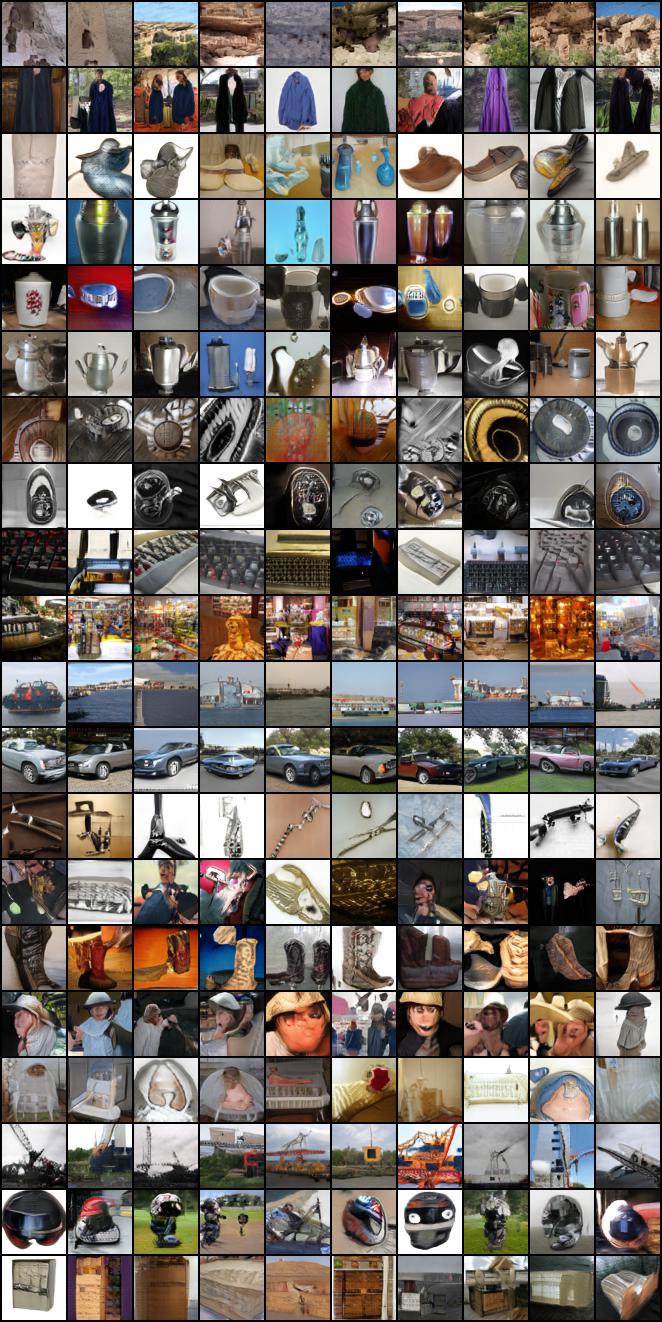}
    \caption{Samples generated using Hinge loss model trained on ImageNet $ 64 \times 64$ for the classes 501-520 (one class per row).}
    \label{fig:I64Baseline2}
\end{figure}

\begin{figure}[ht]
    \centering
    \includegraphics[height=0.8\textheight]{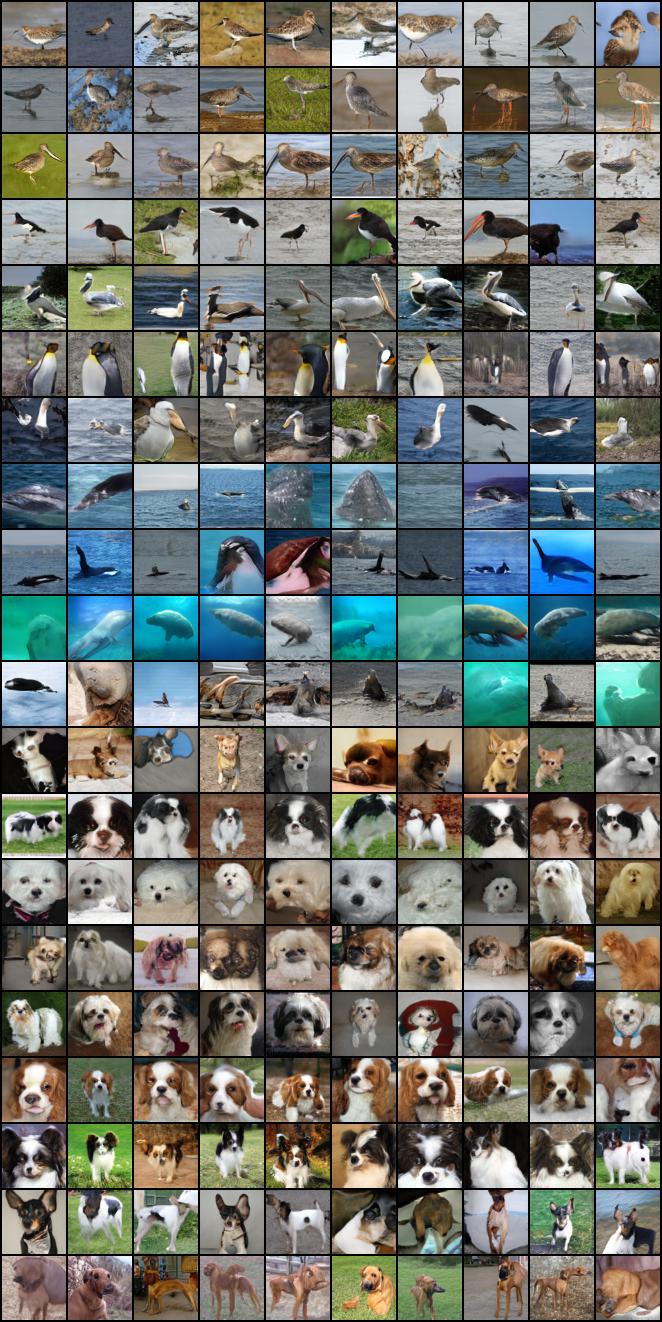}
    \caption{Samples generated using Joint\textdagger \  model trained on ImageNet $ 64 \times 64$ for the classes 141-160 (one class per row).}
    \label{fig:I64Top}
\end{figure}

\begin{figure}[ht]
    \centering
    \includegraphics[height=0.8\textheight]{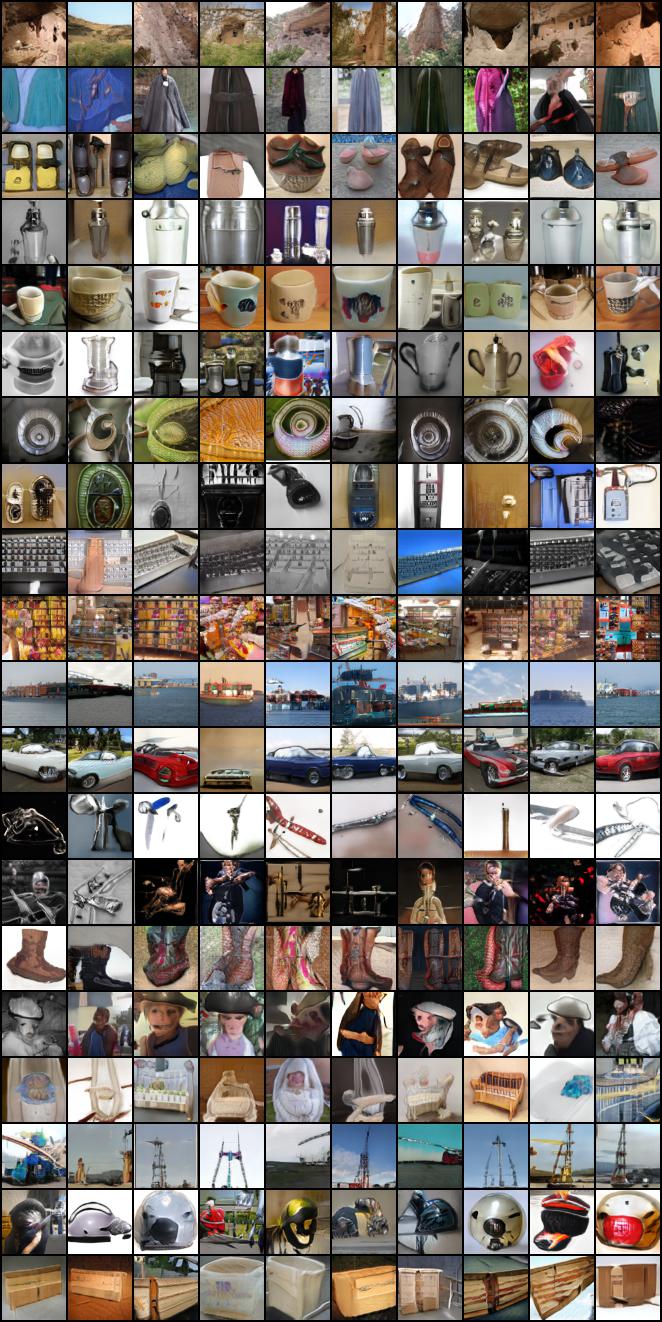}
    \caption{Samples generated using Joint\textdagger \  model trained on ImageNet $ 64 \times 64$ for the classes 501-520 (one class per row).}
    \label{fig:I64Top}
\end{figure}

\end{document}